\documentclass[runningheads]{llncs}
\usepackage{graphicx}
\usepackage{xcolor}
\usepackage{amsmath}
\usepackage{comment}
\begin{document}
\title{VisualWordGrid: Information Extraction From Scanned Documents Using A Multimodal Approach}
\titlerunning{VisualWordGrid}
%
\author{KERROUMI Mohamed\orcidID{0000-0002-2513-8837} \and
SAYEM Othmane\orcidID{0000-0003-0311-9384} \and
SHABOU Aymen\orcidID{0000-0001-8933-7053}}
\authorrunning{Kerroumi et al.}
%
\institute{DataLab Groupe, Credit Agricole S.A, Montrouge , France
\email{kerroumimohamed1997@gmail.com} \\ \email{\{othmane.sayem,aymen.shabou\}@credit-agricole-sa.fr}}
\maketitle              
\begin{abstract}
We introduce a novel approach for scanned document representation to perform field extraction. 
It allows the simultaneous encoding of the textual, visual and layout information in a  3-axis tensor used as an input to a segmentation model. We improve the recent Chargrid and Wordgrid \cite{chargrid} models in  several ways, first by taking into account the visual modality, then by boosting its robustness in regards to small datasets while keeping the inference time low.
Our approach is tested on public and private document-image datasets, showing higher performances compared to the recent state-of-the-art methods.

\end{abstract}

\section{Introduction}
In a vast majority of business workflows, information extraction from templatic documents such as invoices, receipts, tax notices, etc. is largely a manual task. 
Automating this process has become a necessity as the number of client documents increases exponentially.
Most industrial automated systems today have a rule-based approach to documents with a certain structure, and can be associated with a finite number of templates.
However, documents often have  a variety of layouts and structures. In order to understand the semantic content of these documents, the human brain uses the document's layout, as well as the textual and visual information available in its contents.

The challenge is to overcome rule-based systems, and to design end-to-end models that automatically understand both the visual structure of the document and the textual information it contains. For instance, in a document like an invoice, the \textit{total amount to pay} is associated with a numerical value that appears frequently near terms such as \textit{total}, \textit{total to pay} and \textit{net to pay}, and also after fields like \textit{total before taxes}, \textit{taxes}, \textit{cost}, etc. Thus, as Katti et al. showed with Chargrid \cite{chargrid}, combining both positional and textual information, was proven to be efficient for this task. 

On the other hand, the visual content of a document was proven to improve  model accuracy for document classification when combined with textual information \cite{kuider}.

In this article, we prove that adding visual information to textual and positional features improves the performance of the information extraction task.
The improvement is more significant when dealing with documents with rich visual characteristics such as tables, logos, signatures, etc. 
We extend the work of Katti et al. \cite{chargrid,bertgrid} with a new approach (called \textbf{VisualWordGrid}) that combines the aforementioned modalities with two different strategies to achieve the best results in the task of information extraction from image documents.

The present paper is organized as follows: Section 2 presents related work for information extraction. Section 3 describes the datasets we used for evaluation. Section 4 introduces the proposed approach. Section 5 discusses the obtained results. Finally, Section 6 provides our conclusions regarding the new method.

\section{Related work}
Interest in solving the information extraction task has grown in fields where machine learning is used, from Natural Language Processing (NLP) to Computer Vision (CV) domains. Depending on the representation of the document, different methods are applied to achieve the best possible performance. 

For instance, NLP methods transform each document to a 1D sequence of tokens, before applying named entity recognition models to recognize the class of each word \cite{ner}. These methods can be successful when applied to documents with simple layout, such as books or articles. However, for documents like invoices, receipts or tax notices, where visual objects such as tables and grids are more common, these textual methods are less efficient. 
In such cases, structural and visual information are essential to achieve good  performance.

Alternatively, computer vision methods can also be very efficient for this task, specifically for documents like Identity Cards which are very rich with visual features. In these approaches, only the image of the scanned document is given as an input. Object detection and semantic segmentation are among the most used techniques for field extraction \cite{cnn} from these documents. The OCR engine is applied at the end of the pipeline on image crops to extract the text of detected fields. These approaches can be very useful when dealing with documents with normalized templates. For documents with various templates and layouts, these models do not perform  well. 

Most recent studies try to exploit the textual and the layout aspects of the document by combining both NLP and CV methods in the extraction task. In the Chargrid \cite{chargrid} or BertGrid \cite{bertgrid} papers, a document is presented as a 2D grid of characters (or words) embeddings. The idea behind this representation is to preserve structural and positional information, while exploiting textual information contained in the document. Both papers reported significant increase in the performance of information extraction task compared to purely textual approaches. In a more general approach, Zhang et al. \cite{trie} recently proposed TRIE, an end-to-end text reading and information extraction approach, where a module of text reading is introduced. The model mixes textual and visual features of text reading and information extraction to reinforce both tasks mutually, in a multitask approach. 

More recently, Yiheng et al. proposed the LayoutLM \cite{layoutLM}, a new method to leverage the visual information of a document in the learning process. 
Instead of having  the text embedding of each token as the sole input, relative position of tokens in the image and the corresponding feature map of the image crop within the document were added too. Inspired by the BERT model \cite{bert}, Yiheng et al. used  scanned document classification task as a supervised pre-training step for LayoutLM to learn the interactions between text and layout information. Then they enforced this learning by a semi-supervised pre-training using Masked Visual-Lanquage Model (MVLM) as a multi-task learning. The dataset used for pre-training contains 11M documents and the pre-training took 170 hours on 8 GPUs. Hence this approach needs large computational resources.

Other works, focused on solving the document semantic segmentation task, introduced the idea of simultaneously encoding visual, textual and structural modalities into a tensor representation. For instance, Yang et al. \cite{yang2017learning} proposed a multimodal approach to extract the semantic structure of documents using a fully convolutional neural network. Barman et al. \cite{Barman}  proposed a multimodal segmentation model that combines both visual and textual features to extract semantic structure of historical documents.

Compared to these related works, we propose in this paper two multimodal document representation strategies suited to the information extraction task. The first one  is simpler, yet highly effective compared to state-of-the-art multimodal approaches (while improving the extraction scores). The second one is similar to the related works \cite{yang2017learning,Barman} but slightly improved and adapted to the field extraction task (i.e. extracting small text regions).

\section{Data}

In order to evaluate our work, we will show our experiments on two datasets showing interesting visual structures that help improving the information extraction task using multimodal strategies.

\subsubsection{\textbf{RVL-CDIP Dataset \cite{rvl}}} 
\begin{figure}[!h]
\begin{minipage}[t]{0.5\linewidth}
    \centering
    {%
    \setlength{\fboxsep}{0pt}%
    \setlength{\fboxrule}{1pt}%
    \fbox{\includegraphics[width=1\textwidth]{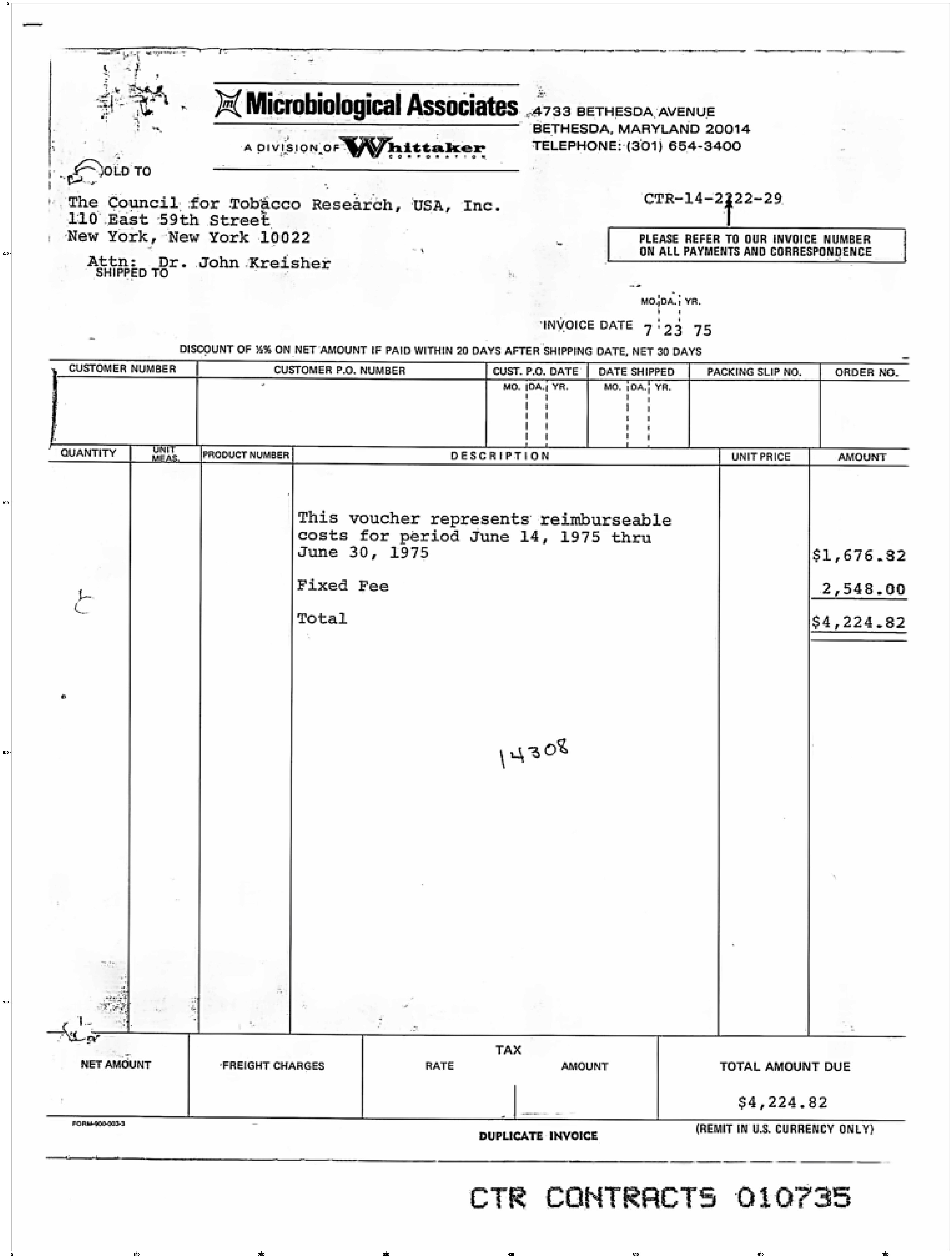}}%
    }
    \label{f31}
\end{minipage}
\hspace{0.1cm}
\begin{minipage}[t]{0.5\linewidth} 
    \centering
    {%
    \setlength{\fboxsep}{0pt}%
    \setlength{\fboxrule}{1pt}%
    \fbox{\includegraphics[width=1\textwidth]{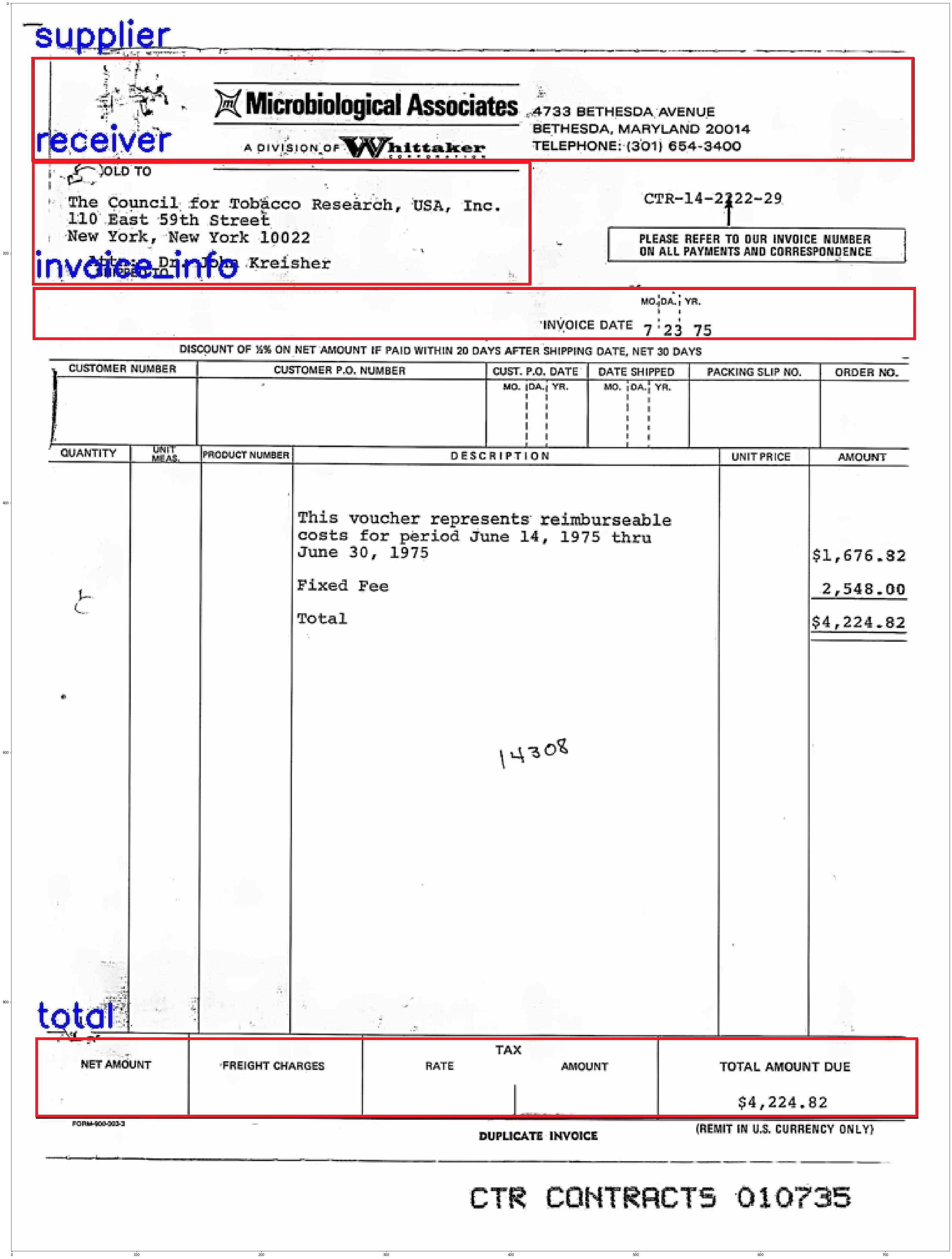}}%
    }%
    \label{f32}
\end{minipage}        
    \caption{Invoice example from RVL-CDIP with fields of interest.}
\label{fig:rvl-cdip}
\end{figure}

It is a public dataset that was released to help improve and evaluate layout analysis techniques on scanned invoice documents. It contains 520 invoice images (Fig.\ref{fig:rvl-cdip}) with their corresponding OCR files containing the extracted text, along with  XML files containing the ground-truth bounding boxes of all the semantic fields. Each word in a given document of the dataset is classified into a semantic region described by a box. Among the 6 available fields, we will focus on extracting 4 of them: \textit{Receiver}, \textit{Supplier}, \textit{Invoice\_info}, \textit{Total}.
\newline

\subsubsection{\textbf{Tax Notice Dataset}}
\begin{figure}[!h]
\begin{minipage}[t]{0.5\linewidth}
    \centering
     {%
    \setlength{\fboxsep}{0pt}%
    \setlength{\fboxrule}{1pt}%
    \fbox{\includegraphics[width=1\textwidth]{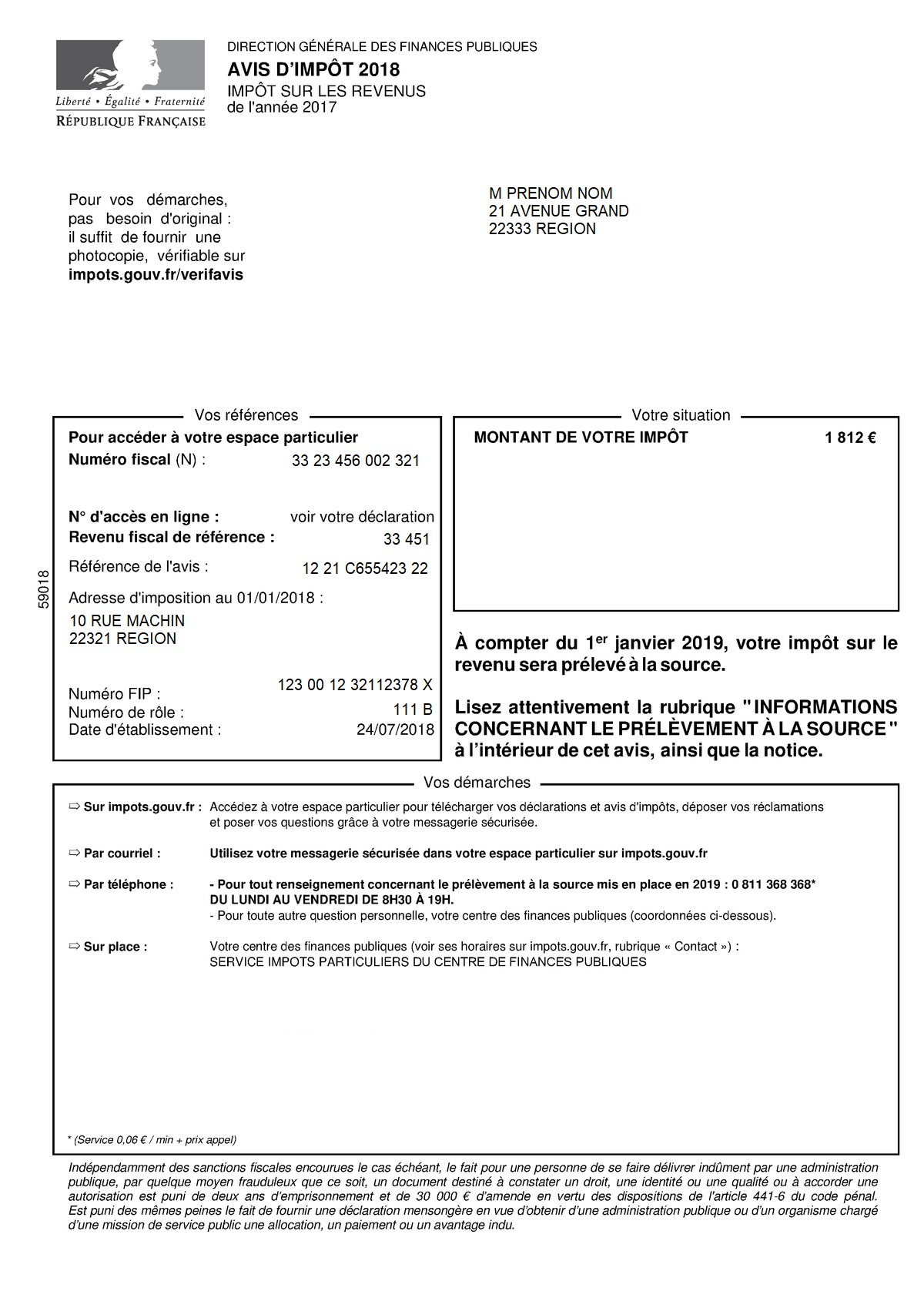}}%
    }%
    \label{f21}
\end{minipage}
\hspace{0.1cm}
\begin{minipage}[t]{0.5\linewidth} 
    \centering
     {%
    \setlength{\fboxsep}{0pt}%
    \setlength{\fboxrule}{1pt}%
    \fbox{\includegraphics[width=1\textwidth]{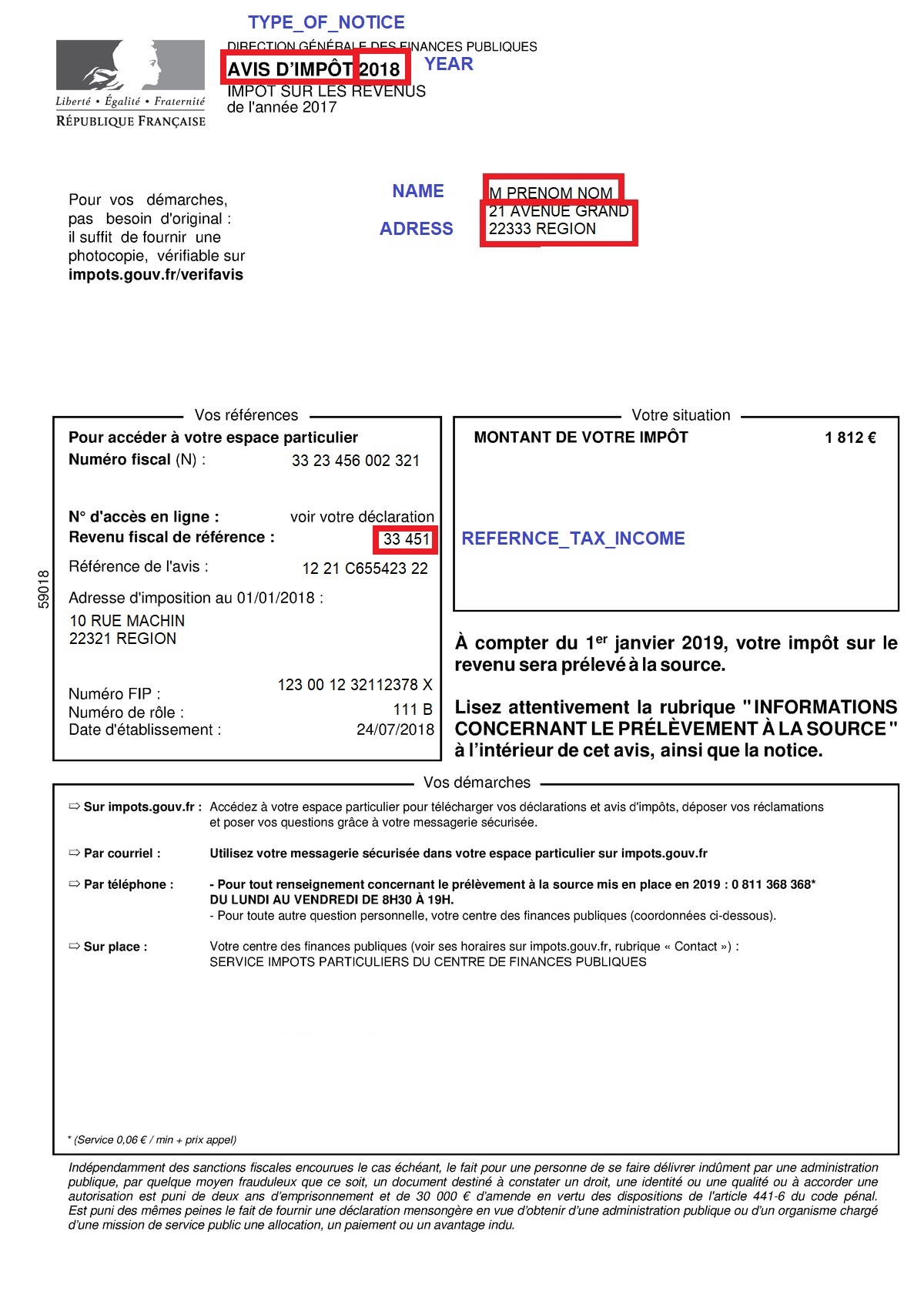}}%
    }%
    \label{f2}
\end{minipage}     
\caption{Tax Notice \textit{fake} example with fields of interest.}
\label{fig:tax}
\end{figure}

It is an in-house \textbf{private} dataset. It contains 3455 tax notices since 2015 (Fig.\ref{fig:tax}). The documents are in French and their templates changed over the years. 
Hence template matching could not be used as an approach for information extraction. The dataset was annotated by manually putting bounding boxes around fields of interest. There are mainly 6 entities to extract from each document: \textit{Year, Name, Address, Type\_of\_Notice, Reference\_Tax\_Income, Family\_Quotient}.
The dataset contains first and second pages from tax notices, as some fields can appear on both pages, depending on the issue date. Moreover, a single page doesn't necessarily contain all fields.

\section{Method}

In this section, we introduce the \textbf{VisualWordGrid} approach, a new 2D representation of documents that extends the Chargrid philosophy by adding the visual aspect of the document to the textual and layout ones. We define two main models that differ on document representation and model architecture : \textbf{VisualWordGrid-pad} and \textbf{VisualWordGrid-2encoders}.

\subsection{Document representation}
Our main idea is adding the visual information of the image to the textual and structural data used in the \textit{WordGrid} representation. The most direct way for doing so is by adding the corresponding RGB channels to each pixel embedding. While this concatenation has no impact on background pixels, it adds a large amount of noise to pre-trained word embeddings. The main challenge here is to adapt the concatenation method to preserve textual embeddings, while adding the background visual information. Our representations of documents extend \cite{chargrid} using two strategies as follows.

Using an OCR, each document can be represented as a set of words and their corresponding bounding boxes. The textual and layout information of each document can be represented in  $ \mathcal{D} = \{(t_k,b_k) | k= 1,...,n\}$, with $t_k$ the k-th token in the text of the document and $b_k = (x_k, y_k, w_k, h_k)$ its corresponding bounding box in the image.
\subsubsection{VisualWordGrid-pad}\hfill

Our first model representation of the document is defined as follows :
 \begin{equation}
    W_{ij}=\left\{\begin{array}{rcl}
    & (e_d(t_k), 0, 0, 0 \big) \quad & if \quad \exists k \quad \text{such} \quad \text{as} \quad (i,j) \prec b_k \\
    & (0_d, R_{ij},G_{ij}, B_{ij}\big)  \quad & \text{otherwise}
\end{array}\right.
\label{eq1}
\end{equation}

$$ (i,j) \prec b_k  \iff   x_k \leq i \leq x_k + w_k \quad  \wedge \quad  y_k \leq j \leq y_k + h_k$$
where $d$ is the embedding dimension, $e_d$ is the  word's embedding function, $0_d$ denotes an all-zero vector of size $d$, and $\big(R_{ij},G_{ij}, B_{ij}\big)$ the RGB channels of the $(i,j)^{th}$ pixel in the raw document's image. 

\begin{figure}[!h]
\centering
\includegraphics[width=\textwidth]{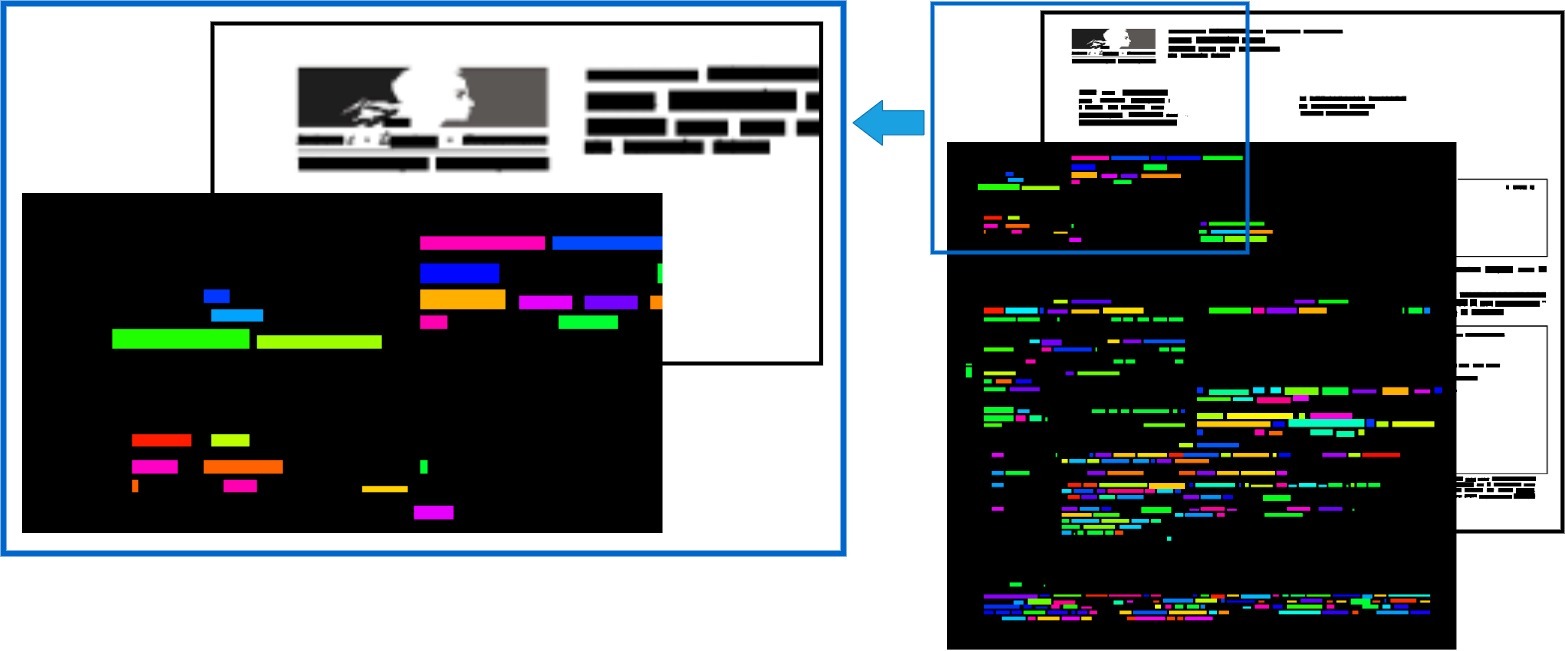}
\caption{VisualWordGrid encoding of an invoice sample. On the right, the proposed concatenation of a Wordgrid representation  and the image of the document. On the left, a zoom of the previous figure.}
\label{fig:visualwordgrid}  
\end{figure}

In other words, for each point in the document's image, if this point is included in a word's bounding box $b_k = (x_k, y_k, w_k, h_k)$, the vector representing this point is the word's embedding padded by $\mathbf{0}_3$. Thus, by setting the RGB channels to $\mathbf{0}_3$, we drop the visual information related to this point. However if the point is not included in any word's bounding box, the vector representing this point is the concatenation of $\mathbf{0}_d$ and the RGB channels of this point. In this case, we keep the visual information.
Hence, the visual, textual and layout information of the document are encoded simultaneously in a 3-axis tensor of shape $( H, W, d+3)$ as shown in Fig.\ref{fig:visualwordgrid}, while preserving their original information.

\subsubsection{VisualWordGrid-2encoders}\hfill

Our second model representation is similar to the CharGrid-Hybrid approach presented in \cite{chargrid}. Instead of encoding the document on the character level using a one-hot encoding, we encode the document on the word level using \textit{Word2Vec} \cite{word2vec}  or  \textit{Fasttext} \cite{fasttext} embeddings. Hence, for each document we have two inputs:
\begin{itemize}
\item \textit{WordGrid encoding}: This input encodes the textual and  layout information of the document.  This approach of encoding is similar to WordGrid presented in \cite{chargrid}.
For words encoding, we use \textit{Word2Vec} or  \textit{Fasttext} embeddings.

\begin{equation}\label{eq3}
   W_{ij}=\left\{\begin{array}{rcl}
                     & e_d(r_k) & \quad if \quad \exists k \quad \text{such}\quad \text{as} \quad (i,j) \prec b_k \\
                     & 0_d   \quad  & \quad \text{otherwise}
\end{array}\right.  
\end{equation}

\item \textit{Image}: The raw image of the document resized to match the WordGrid encoding dimensions. 
\end{itemize}

\subsection{Model Architectures}

In this section, we discuss model architectures related to both strategies.

\subsubsection{VisualWordGrid-pad}\hfill

Once the 2D representation of the document is encoded, we use it to train a neural segmentation model. Unlike \textit{chargrid} and \textit{wordgrid} papers, we dropped the bounding box regression block to keep the semantic segmentation block only, since there can be at most one instance of each class in the datasets. 

We use the \textit{Unet} \cite{unet} as a segmentation model and the \textit{ResNet34} \cite{resnet} as a backbone for the encoder. The weights of the backbone are initialized using transfer learning from a model pre-trained on the ImageNet classification task. These weights are available in the open-source package \textit{Segmentation Models} \cite{sm}. The UNet component extracts and encodes advanced features of the input grid in a small feature map, and the decoder expands this feature map to recover segmentation maps of the same size as the input grid, and thus generates the predicted label masks. We used a \textit{softmax} activation function for the final layer of the decoder. The shape of the decoder's output is $(H,W,K+1)$, where $K$ is the number of fields of interest, and $1$ is the background class. 

\begin{figure*}
  \includegraphics[width=\textwidth, height=\textheight]{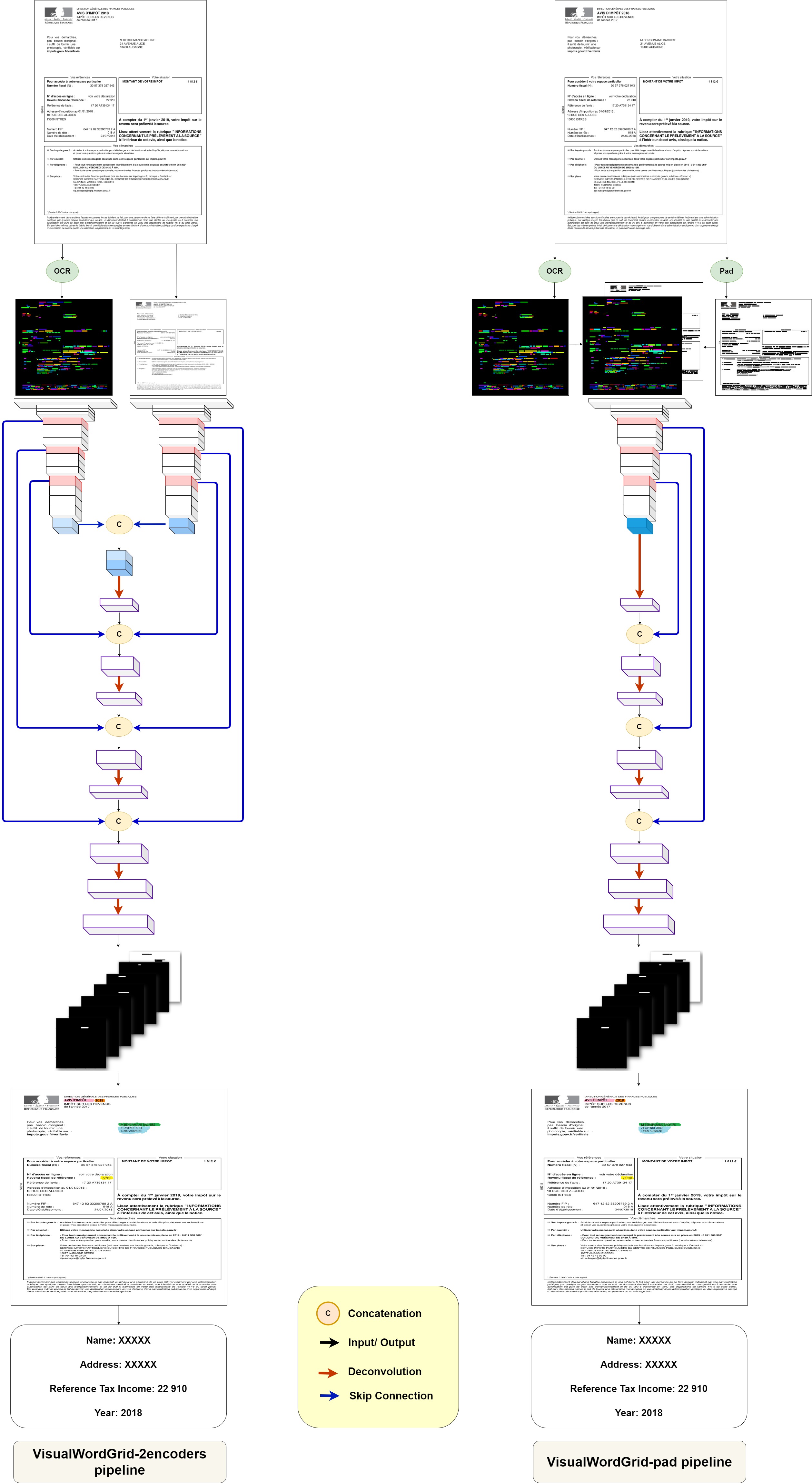}
  \caption{VisualWordGrid pipelines.}
  \label{visualwordgrid}
\end{figure*}

In the inference step, we iterate over the bounding box of each word in the OCR output, then attribute a single class to the most dominant category pixel-wise, to get the final prediction value for the corresponding field. (see Fig.\ref{visualwordgrid}).

\subsubsection{VisualWordGrid-2encoders}\hfill

This model is composed of 2 encoders. The first one is the classic WordGrid encoder \eqref{eq3}, and the second one is the raw image encoder. We keep one decoder, and for each block in the decoder, we concatenate on the skip connections from both encoders (see Fig.\ref{visualwordgrid}).

\subsection{Implementation Details}

In this section, we provide implementation details.

\subsubsection{Word embedding function}\hfill

We use different word embedding functions depending on the dataset. For RVL-CDIP, we propose Word2Vec pretrained embeddings on the Wikipedia corpus, publicly available thanks to \textit{Wiki2Vec} \cite{wiki2vec}. The choice of \textit{Wiki2Vec} is due to the good quality of the OCR files, since words are correctly recognized and most of them have their related embeddings. Unlike RVL-CDIP dataset, OCR outputs of our Tax Notice dataset are noisy, due to the quality of customer documents scans. We observed frequent misspelling errors in the Tesseract 4.1 \cite{tesseract} ouputs. Our experiments show that a custom \textit{ FastText} embedding trained on the corpus of the Tax Notice dataset is the best words embedding function to handle the noise. As explained in \cite{fasttext}, Word embedding using this approach is the sum of n-grams subword embeddings. Hence, even in case of a misspelled or  dropped character in the token, its embedding wouldn't differ too much from the embedding of the original word.

\subsubsection{Loss function}\hfill

The loss function we use for training is the sum of the cross entropy loss for segmentation $(L_{seg})$ and the Intersection over Union loss $(L_{IoU})$.

\begin{equation}
   Loss = L_{seg} + L_{IoU}
   \label{eq4}
\end{equation}

\begin{itemize}
\item \textbf{Cross Entropy Loss}: The cross entropy loss penalizes pixel mis-classification. The goal is to assign each pixel to  its ground truth field.
\begin{equation}
    L_{seg} = \sum_{x\in \Omega } -log\big (p_{\hat l(x)}(x)\big )
    \label{eq7}
\end{equation}

with $\hat l(x)$ the ground truth label of the point $x$.\\

\item \textbf{Intersection over Union Loss}: This loss function is often used to train neural networks for a segmentation task. It's a differentiable approximation of the IoU metric and is the most indicative of success for segmentation tasks as explained in \cite{IoU}. In our case, it significantly increases performances of the model compared to a model trained only with the cross entropy. The IoU metric is defined as : \\
\begin{equation}
    IoU = \frac{I}{U} = \frac{|T\cap P|}{|T\cup P|} = \frac{|T\odot P|}{|T+P-(T\odot P)|} 
    \label{eq8}
\end{equation}
\begin{equation}
   L_{IoU} = 1 - IoU 
    \label{eq9}
\end{equation}
where $T$ is the true labels of the image pixels and $P$ is their prediction labels. We also use the $IoU$ metric to monitor the training of our models.
\end{itemize}

\subsubsection{Metrics}\hfill

To evaluate the performance of the different models, we used two metrics:
\begin{itemize}
    \item \textbf{Word Accuracy Rate (WAR)}: It's the same metric as the one used in \cite{chargrid}. It's similar to the \textit{Levenshtein distance}  computed on the token level instead of the character level. It counts the number of substitutions, insertions and deletions between the ground-truth and the predicted instances. This metric is usually used to evaluate speech-to-text models. The WAR formula is as follows:
    \begin{equation}
        WAR = 1 - \frac{\#[insertions] + \#[deletions] + \#[substitutions] }{N}
        \label{eq17}
    \end{equation}

where, $N$ is the total number of tokens in the ground truth instance for a specific field. The WAR of a document is the average on all fields. 

 \item \textbf{Field Accuracy Rate (FAR)}: This metric evaluates the performance of the model in extracting  complete and exact field information. A field is the set of words of a same entity. This metric counts the number of exact match between the ground-truth and the predicted instances. It is useful in industrial applications, as we need to evaluate the number of cases where the model succeeds to extract the whole field correctly, for control purposes for example. The FAR formula is as follows:
 \begin{equation}
FAR = \frac{ \#[\text{Fields exact Match}]}{N_{fields}}
    \label{eq19}
 \end{equation}
where, $\#[\text{Fields exact Match}]$ is the number of fields correctly extracted from the document with an exact match between  the ground-truth and the predicted words values, and $N_{fields}$ is the total number of fields. We note that for any processed document in the evaluation set, with no target field in the ground truth, we attribute an empty string to the value of each field, so false positives are  penalized too.\\
In the next section, we will report for each model the average WAR and FAR metrics on the documents in the test set.
\end{itemize}

\section{Experiments}
In this section, we compare our approaches (VisualWordGrid-pad, VisualWordGrid-2encoders) to two others, on both datasets. We report their average scores ($\overline{FAR}$, $\overline{WAR}$) and inference time ($\overline{InferenceTime}$)  on CPU for a single document, and their number of trainable parameters.

The two competing approaches are the following ones:
\begin{itemize}
    \item \textbf{Layout Approach}: This approach is a layout encoding only. Instead of using word embedding or pixel RGB channels to encode a specific document as in \eqref{eq1}, it uses a simpler 2D encoding suited to a segmentation task, i.e.:
 \begin{equation}
    W_{ij}=\left\{\begin{array}{rcl}
    & ( 1, 1, 1 \big) \quad & if \quad \exists k \quad \text{such} \quad \text{as} \quad (i,j) \prec b_k \\
    & (0, 0, 0\big)   \quad & \text{otherwise}
\end{array}\right.
\label{eq10}
\end{equation}

\begin{figure}[!h]
\begin{minipage}[t]{0.5\linewidth}
    \centering
    {%
    \setlength{\fboxsep}{0pt}%
    \setlength{\fboxrule}{1pt}%
    \fbox{\includegraphics[width=1\textwidth]{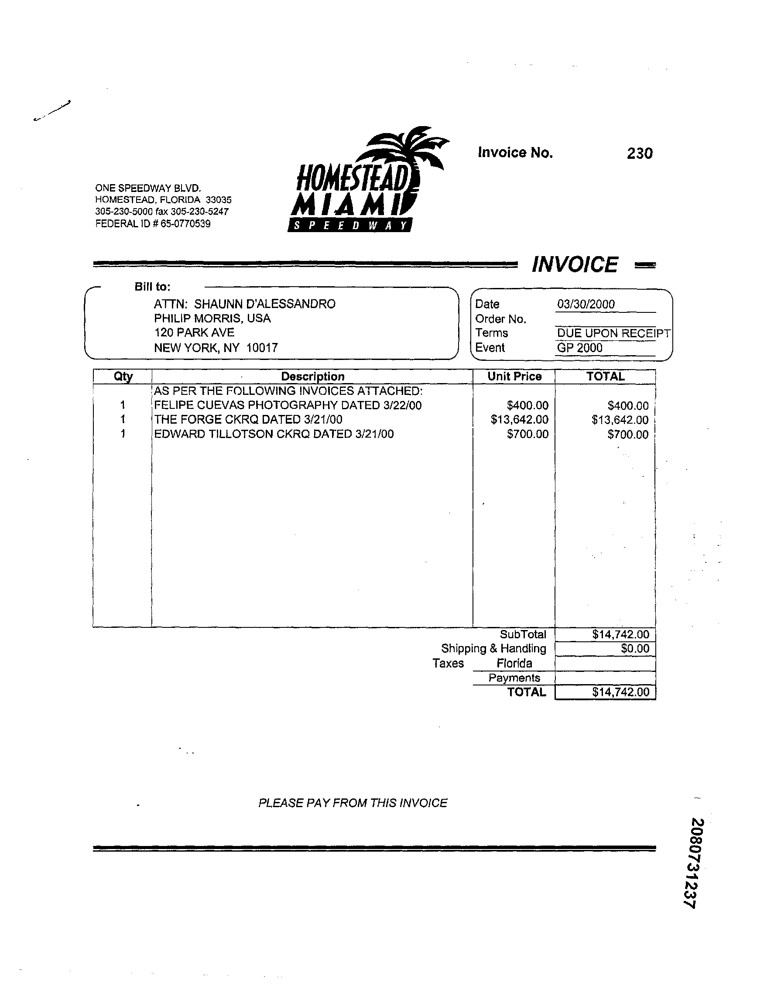}}%
    }%
    \label{f11}
\end{minipage}
\hspace{0.1cm}
\begin{minipage}[t]{0.5\linewidth} 
    \centering
     {%
    \setlength{\fboxsep}{0pt}%
    \setlength{\fboxrule}{1pt}%
    \fbox{\includegraphics[width=1\textwidth]{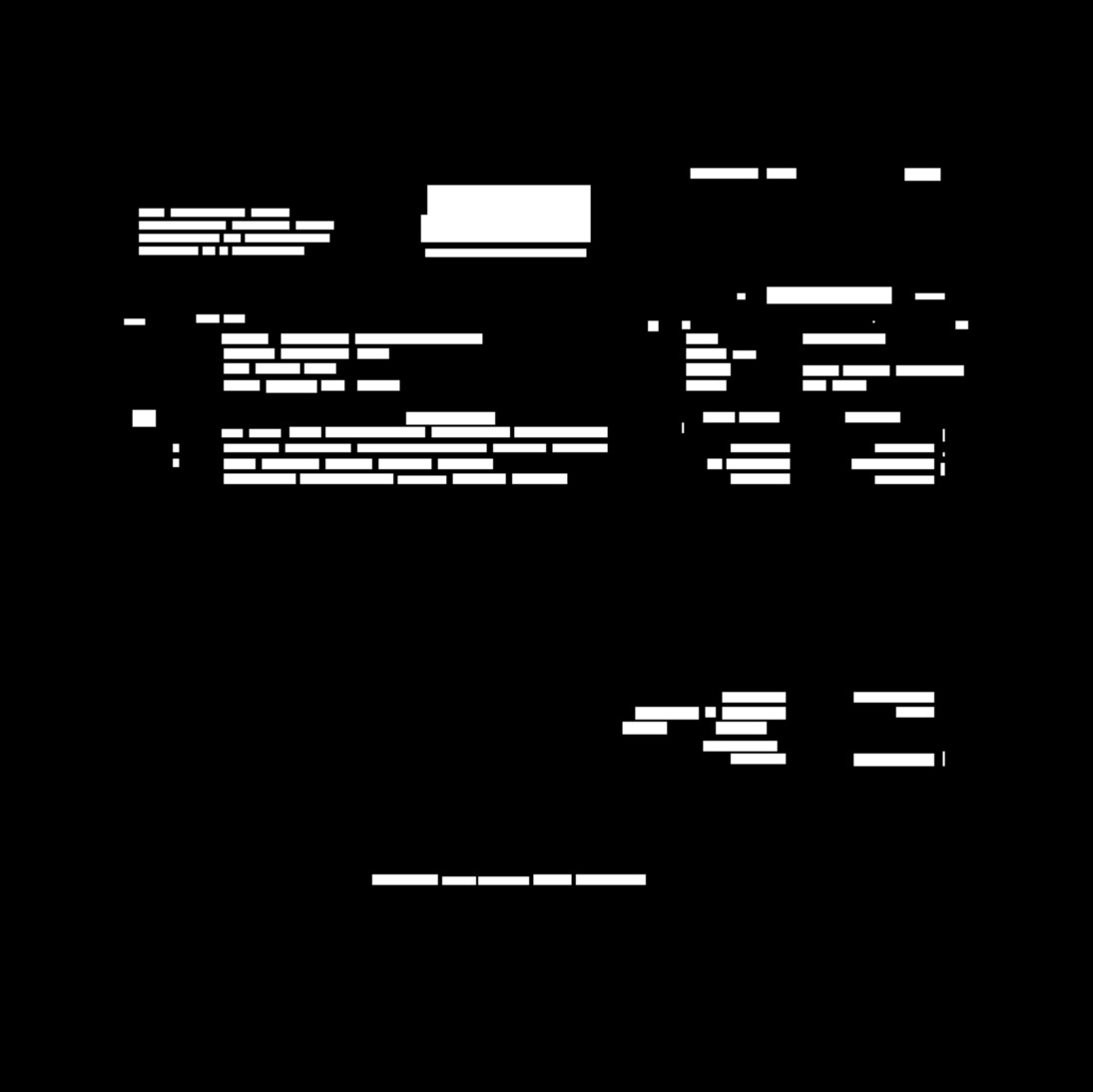}}%
    }%
    \label{f22}
\end{minipage}        
\caption{Invoice sample and its encoding using the Layout approach}
\label{fig:layout}
\end{figure}

Then, we use this type of document encoding (Fig. \ref{fig:layout}) as input to train an information extraction model  using the proposed architecture, loss and model hyper-parameters. 

\item \textbf{WordGrid}: This approach is very similar to BertGrid \cite{bertgrid}. Instead of using a Bert \cite{bert} model to generate contextual embeddings, we use a \textit{Word2Vec} pre-trained embedding for RVL-CDIP dataset, and  a custom \textit{Fasttext} embedding for the Tax Notice dataset. Equation \eqref{eq3} introduces the document encoding formula. 

We keep the same model architecture, loss function and model hyper-parameters as  proposed in the VisualWordGrid model.


\end{itemize}
\subsection{Datasets}
Since the RVL-CDIP dataset volume is very small, we don't use a classic split of the dataset into training set and validation sets. Instead,  we use a \textit{k-fold} split of the dataset with $k=5$. For each experiment, we do 5 tests, each one with a training on 80\% of the dataset and the remaining 20\% is split equally into validation and test sets. We report the average of the metrics on the 5 tests. This way, the values of the metrics don't depend on the seed of the split, and metrics are a more reliable representation of real model performance.  

For the Tax Notice dataset, we assign 80\% of the dataset to training, 15\% to validation and 5\% to test, on which we report our results.  The OCR task to extract textual information was performed using the open source OCR engine Tesseract 4.1.

 \subsection{Results}
For all experiments , we use \textit{Adam} optimizer with \textit{lr=0.001} and \textit{batch\_size}=8. We use a GPU NVIDIA Quadro RTX 6000 with 24GB GPU memory and the  \textit{Keras} framework \cite{keras} to build and train models. The inference time is measured on  Intel Xeon W-2133 CPU (3.60 GHz).  The table \ref{tab1} shows the scores ($\overline{WAR}$, $\overline{FAR}$) of the different approaches on the RVL-CDIP dataset.\\

\begin{table}[!htbp] 
\label{tab:schemes} 
\caption{Models performances on the RVL-CDIP dataset.}
\label{tab1}
\centering 
\renewcommand{\arraystretch}{1.5} 
\begin{tabular}{|c|c|c|c|c|}
\hline
 Approach & $\overline{FAR}$ &  $\overline{WAR}$ & $\overline{Inference Time}$ & \#Parameters   \\
\hline

Layout Only  & 23.0 \%   &   5.4 \% & 2.14 s & 24 439 384\\
\hline
WordGrid  & 27.7 \%   &  10.8 \% & 2.22 s  &  24 743 673 \\
\hline
\textbf{VisualWordGrid-pad}  & \textbf{28.7 \%}   &  \textbf{18.7 \%} &
3.77 s & 
24 753 084\\
\hline
VisualWordGrid-2encoders  & 26.9 \%   &  17.0 \% & 
6.08 s & 48 003 004\\

\hline
\end{tabular}
\end{table}

We clearly see in the table \ref{tab1} that \textbf{VisualWordGrid-pad} gives the best FAR and WAR scores. Our proposed encoding system improves the WordGrid FAR and WAR by \textbf{1 } and  \textbf{7.9 } respectively. Moreover, it exploits all the visual, textual and structural content of documents while keeping the inference time and  the number of  parameters close to the WordGrid ones. 

Unlike Katti et al. \cite{chargrid}, we notice an increase in the WAR score when using the two encoders approach (VisualWordGrid-2encoders)  to capture the visual and textual aspect of document. It boosts the WordGrid performance, since the WAR goes up by \textbf{6.2 }.
The reasons for this improvement are the modifications we added to make the model  more robust in the information extraction task. We used a ResNet34 backbone for the encoder and took advantage of transfer learning to speed up the training of the model. We also changed the cross entropy loss used in \cite{chargrid} by adding the  IoU loss to it. Notice that we  used the $IoU$ as a metric for the callback. 

Similarly, we tested the different approaches on the Tax Notice dataset. We reported the results in table \ref{tab2}.
 \begin{table}[!htbp] 
\label{tab:schemes2} 
\caption{Models performances  on the Tax Notice dataset.}
\label{tab2}
\centering 
\renewcommand{\arraystretch}{1.5} 
\begin{tabular}{|c|c|c|c|c|}
\hline
 Approach & $\overline{FAR}$ &  $\overline{WAR}$ & $\overline{Inference Time}$ & \#Parameters   \\
\hline

Layout Only  & 83.3 \%   &   92.3 \% & 5.29 s & 24 439 674\\
\hline
WordGrid  & 83.6 \%   &  92.4 \% & 5.70 s & 24 743 963\\
\hline
VisualWordGrid-pad  & 83.9 \%  &  92.9 \% & 5.92 s &  24 753 374\\
\hline
\textbf{VisualWordGrid-2encoders}  & \textbf{85.8} \%   &  \textbf{93.6 \%} & 6.19 s & 48 003 294 \\

\hline
\end{tabular}
\end{table}\\

The VisualWordGrid-padding approach slightly improves the WordGrid scores, while the VisualWordGrid-2encoders gives the best performance but at the expense of a slightly  higher inference time.

As in several industrial applications, using information extraction requires the smallest inference time. VisualWordGrid-pad would be the best choice. It leverages the visual/textual/layout information of a document while keeping the number of trainable parameters roughly the same as WordGrid. 

\section{Conclusion}
VisualWordGrid is a simple, yet  effective 2D representation of documents that encodes the textual, layout and visual information simultaneously. The grid-based representation includes token embeddings and the image's RGB channels. We can take advantage of these multimodal inputs to perform several document understanding tasks. For the information extraction task, VisualWordGrid shows better results than those of  state of the art models on two datasets (the public RVL-CDIP dataset and  the private Tax Notice dataset), while keeping model parameters and inference time roughly the same (especially when using the padding strategy). In many fields, this approach is suitable  for production.

\bibliographystyle{splncs04}
\bibliography{VisualWordGrid}


\end{document}